\documentclass{article} % For LaTeX2e
\usepackage{iclr2022_conference,times}

\usepackage{amsmath,amsfonts,bm}

\def\eqref#1{equation~\ref{#1}}
\def\1{\bm{1}}

\DeclareMathAlphabet{\mathsfit}{\encodingdefault}{\sfdefault}{m}{sl}
\SetMathAlphabet{\mathsfit}{bold}{\encodingdefault}{\sfdefault}{bx}{n}

\iclrfinalcopy
\usepackage{hyperref}
\usepackage{url}
\usepackage{graphicx}
\usepackage{amsmath}
\usepackage{wrapfig}
\usepackage{algorithmic}
\usepackage{wrapfig}
\usepackage[linesnumbered,ruled,vlined]{algorithm2e}
\usepackage{comment}
\usepackage{pifont}
\newcommand{\cmark}{\ding{51}}%
\newcommand{\xmark}{\ding{55}}%

\title{A Fast and Efficient Conditional Learning for Tunable Trade-off between Accuracy and Robustness}

\author{Souvik Kundu\thanks{ Work done during his internship at Intel Labs.} \\
Department of Electrical and Computer Engineering\\
University of Southern California\\
Los Angeles, CA 90089, USA \\
\texttt{souvikku@usc.edu} \\
\And
Sairam Sundaresan\\
Intel AI Labs \\
San Diego, CA 92127, USA \\
\texttt{sairam.sundaresan@intel.com} \\
\And
Massoud Pedram\\
Department of Electrical and Computer Engineering\\
University of Southern California\\
Los Angeles, CA 90089, USA \\
\texttt{{pedram}@usc.edu} \\
\And
Peter A. Beerel\\
Department of Electrical and Computer Engineering\\
University of Southern California\\
Los Angeles, CA 90089, USA \\
\texttt{{pabeerel}@usc.edu} \\
}

\begin{document}

\maketitle

\begin{abstract}
Existing models that achieve state-of-the-art (SOTA) performance on both clean and adversarially-perturbed images rely on convolution operations conditioned with feature-wise linear modulation (FiLM) layers. These layers require many new parameters and are hyperparameter sensitive. They significantly increase training time, memory cost, and potential latency which can prove costly for resource-limited or real-time applications. 
In this paper, we present a \textit{fast learnable once-for-all adversarial training} (FLOAT) algorithm, which instead of the existing FiLM-based conditioning, presents a unique \textit{weight conditioned} learning that requires $\textbf{no}$ additional layer, thereby incurring no significant increase in parameter count, training time, or network latency compared to standard adversarial training. In particular, we add configurable scaled noise to the weight tensors that enables a trade-off between clean and adversarial performance. 
Extensive experiments show that FLOAT can yield SOTA performance improving both clean and perturbed image classification by up to $\mathord{\sim}6\%$ and $\mathord{\sim}10\%$, respectively. Moreover, real hardware measurement shows that FLOAT can reduce the training time by up to $1.43\times$ with fewer model parameters of up to $1.47\times$ on iso-hyperparameter settings compared to the FiLM-based alternatives. 
Additionally, to further improve memory efficiency we 
introduce FLOAT $\textit{sparse}$ (FLOATS), a form of non-iterative model pruning  and provide detailed empirical analysis to provide a three way accuracy-robustness-complexity trade-off for these new class of pruned conditionally trained models. 
\end{abstract}

\section{Introduction}
\label{sec:intro}
 With the growing usage of DNNs in safety-critical and sensitive applications including autonomous-driving \cite{bojarski2016end} and medical image analysis \cite{han2021advancing}, it has become crucial that they have high classification accuracy on both clean and adversarially-perturbed images \cite{wang2020once}. To improve the DNN model performance against these adversarial samples\footnote{Adversarial images consists of imperceptible pixel perturbations from corresponding clean ones and can fool a well trained classifier models into making wrong predictions.}, various defense mechanisms have been proposed including hiding gradients \cite{tramer2017ensemble}, adding noise to parameters \cite{he2019parametric}, and detection of adversaries \cite{meng2017magnet}. In particular, adversarial training \cite{madry2017towards,hua2021bullettrain} has proven to be a consistently effective approach in achieving state-of-the-art robustness.  

These defenses, however, come at various costs. Firstly, most of these methods suffer from increased training times due to the additional back-propagation overhead caused by generating perturbed images. Secondly, adversarial defenses sometimes cause a significant drop in clean-image accuracy \cite{tsipras2018robustness}, highlighting an accuracy-robustness trade-off that has been explored both theoretically and experimentally \cite{sun2019towards}, \cite{tsipras2018robustness}, \cite{schmidt2018adversarially}. Moreover, the defenses rely on several hyperparameters whose settings force the model to work at a specific point along this trade-off. This is disadvantageous in applications in which the desired trade-off depends on context \cite{wang2020once}. 

A naive solution to this problem is to use multiple networks trained with different priorities between clean and adversarial images. This however, comes with the heavy cost of both increased training time and inference memory. Alternatively, recent work has proposed training a once-for-all adversarial network (OAT) that supports \textit{conditional learning} \cite{wang2020once}, enabling the network to adjust to different input distributions.
In particular, after each batch-normalization (BN) layer, they add a feature-wise linear modulation (FiLM) module \cite{perez2018film} whose weights are controlled by a parameter $\lambda$. For inference, the user sets $\lambda$ to enable an in-situ trade-off between accuracy and robustness. The disadvantage with this approach is that the added FiLM modules increase the parameter count, training time, and network latency, limiting applicability in resource-constrained, real-time applications. Moreover, our investigation shows that the CA-RA performance of OAT is heavily dependent on the choice of training hyperparameter $\lambda$s. For example, the accuracy with ResNet34 on CIFAR-10 varies up to $21.97\%$.
\begin{figure}[!t]
    \includegraphics[width=0.88\textwidth]{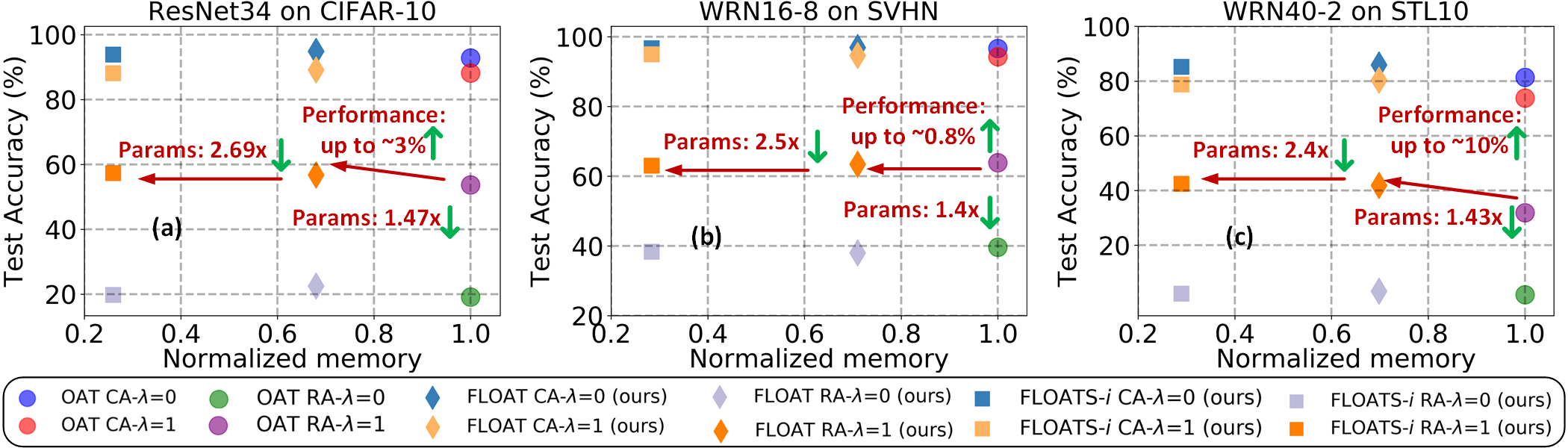}
  \centering
  \vspace{-2mm}
  \caption{Normalized memory vs. Test accuracy for FLOAT and FLOAT with irregular sparsity (FLOATS-$i$) compared to the existing state-of-the-art OAT  for (a) ResNet34, (b) WRN16-8, and (c) WRN40-2, respectively. CA and RA represent clean-image classification accuracy and robust accuracy (accuracy on adversarial images), respectively. For each model we normalized the memory requirement with the maximum memory needed to store corresponding model.}
  \label{fig:float_vs_oat_intro_perf}
  \vspace{-7mm}
\end{figure}

\textbf{Our contributions.} In this paper, our contributions are two-fold.
First, in view of the above concerns, we present a \textit{fast learnable once-for-all adversarial training} (FLOAT). In FLOAT, we train a model using a novel mechanism wherein each weight tensor of the model is transformed by conditionally adding a noise tensor based on a binary parameter $\lambda$, yielding state-of-the-art (SOTA) test accuracy for clean and adversarial images by in-situ setting $\lambda=0.0$ and $1.0$, respectively.
For inference, we further show that model robustness can be correlated to the strength of the noise-tensor scaling factor. This motivates a simple yet effective noise re-scaling approach controlled by an user-provided floating-point parameter that can help the user to have a practical accuracy-robustness trade-off. Because FLOAT does not require additional layers to perform conditioning, it incurs no increase in latency and causes only a negligible increase in parameter count compared to the baseline models. Moreover, compared to OAT, FLOAT training is up to $1.43\times$ faster, attributable to the fact that FLOAT does not  require training with intermediate fine-grained values of $\lambda$s. 

%ECCV new
Secondly, for efficient deployment of the models to resource-limited edge devices, we present FLOAT sparse (FLOATS), an extension of FLOAT, that not only provides adaptive tuning between RA and CA, but also facilitates high levels of model compression (via pruning) without incurring any additional training time. In particular, we propose and empirically evaluate the efficacy of FLOATS with both irregular and structured channel pruning, namely FLOATS-$i$ and FLOATS-$c$, respectively.  However, despite the potential speed-up on underlying hardware \cite{liu2018rethinking}, channel pruning often costs classification performance \cite{kundu2021dnr} because of its strictly constrained form of sparsity. 
We thus extend FLOATS to propose a globally-structured locally-irregular hybrid sparsity. In particular, we perform channel reduction through network slimming \cite{yu2018slimmable} reducing latency and memory usage, and use irregular pruning in conjunction with this to further reduce memory cost.
%can yield both  while performing irregular pruning on the original model. 
These new models not only provide compression, but enable an in-situ inference trade-off across accuracy, robustness, and complexity.
%support accuracy-robustness trade-off on networks with different slimming conditions. 
%

To evaluate the merits of FLOAT, we conduct extensive experiments on CIFAR-10, CIFAR-100, Tiny-ImageNet, SVHN, and STL10 with ResNet34 (\textcolor{black}{on both CIFAR and Tiny-ImageNet datasets}), WRN16-8, WRN40-2, respectively. As shown in Fig. \ref{fig:float_vs_oat_intro_perf}, compared to OAT, FLOAT can provide improved accuracies of up to $\mathord{\sim}6\%$, and $\mathord{\sim}10\%$, on clean and perturbed images, respectively, with reduced parameter budgets of up to $1.47\times$. FLOATS can yield even further parameter-efficiency of up to $2.69\times$ with similar CA-RA benefits.

%The remainder of this paper is organized as follows. Section \ref{sec:back} presents the necessary preliminaries and prior work. We present our approach in Section \ref{sec:approach} and analyze experimental results in Section \ref{sec:expt}.  Finally, the paper concludes in Section \ref{sec:conc} with discussions of broader impact in Section \ref{sec:broad}.  

\section{Preliminaries}
\label{sec:back}
\subsection{Notation}

Consider a model $\Phi$ with $L$ layers parameterized by $\bm{\Theta}$ that learns a function $f_{\Phi}(.)$. For a classification task on
dataset $\bm{X}$ with distribution $D$, the model parameters $\bm{\Theta}$ are learned by minimizing the empirical risk (ERM) as follows 
\vspace{-6mm}
\begin{align}
     \mathcal{L}(f_{\Phi}(\bm{x}, \bm{\Theta}; t)),
\end{align}
where $t$ is the ground-truth class label, $\bm{x}$ is the vectorized input drawn from $\bm{X}$, and $\mathcal{L}$ is the cross-entropy loss function.

\subsection{Robust Model Training}

Several forms of adversarial training (AT) have 
been proposed to improve robustness \cite{madry2017towards}, \cite{samangouei2018defense}, \cite{buckman2018thermometer}. They use clean as well as adversarially-perturbed images to train a model.
Projected gradient descent (PGD) attack, recognized as one of the strongest $L_{\infty}$ adversarial example generation algorithms \cite{madry2017towards}, is typically used to create adversarial images during training. The perturbed image for a PGD-$k$ attack with $k$ as the number of steps is given by
\begin{align}
\hat{\bm {x}}^{k}&=\texttt{Proj}_{P_{\epsilon}(\bm{x})} (\hat{{\bm x}}^{k-1} + \sigma \times \texttt{sign}(\nabla_{x}\mathcal{L}(f_{\Phi}(\hat{\bm x}^{k-1}, \bm{\Theta}; t))) 
\label{eq:pgd}
\end{align}
Here, the scalar $\epsilon$ corresponds to the perturbation constraint that determines the severity of the perturbation. $\texttt{Proj}$ projects the updated adversarial sample onto the projection space $P_{\epsilon}(\bm{x})$,  which is the   $\epsilon$-$L_{\infty}$ neighbourhood of the benign sample  $\bm{x}$\footnote{Note that the generated $\hat{\bm {x}}$ are clipped to a valid range which, for our experiments, is $[0,1]$.}. $\sigma$ is the attack step-size. For PGD-AT, the model parameters are then learned by the following ERM 
\begin{align}
     [\underbrace{(1-\lambda)\mathcal{L}(f_{\Phi}(\bm{x}, \bm{\Theta}; t))}_{\mathcal{L}_C} + \underbrace{\lambda\mathcal{L}(f_{\Phi}(\hat{\bm{x}}, \bm{\Theta}; t))}_{\mathcal{L}_A}],
\label{eq:adv_loss}
\end{align}
where $\mathcal{L}_C$ and $\mathcal{L}_A$ correspond to the clean and adversarial image classification loss components, respectively, weighted by the scalar $\lambda$. Hence, for a fixed $\lambda$ and adversarial strength, the model learns a fixed tradeoff between accuracy and robustness. For example, an AT with $\lambda$ value of $1$ will allow the model to completely focus on perturbed images, resulting in a significant drop in clean-image classification accuracy. \textcolor{black}{Another strategy to improve model robustness is through the addition of noise to the model weight tensors. For example, \cite{he2019parametric} introduced the idea of noisy weight tensors with a learnable noise scaling factor and improved robustness against gradient-based attacks. However, this strategy also incurs a significant drop in clean image classification accuracy.}
\begin{figure}[!t]
    \includegraphics[width=0.80\textwidth]{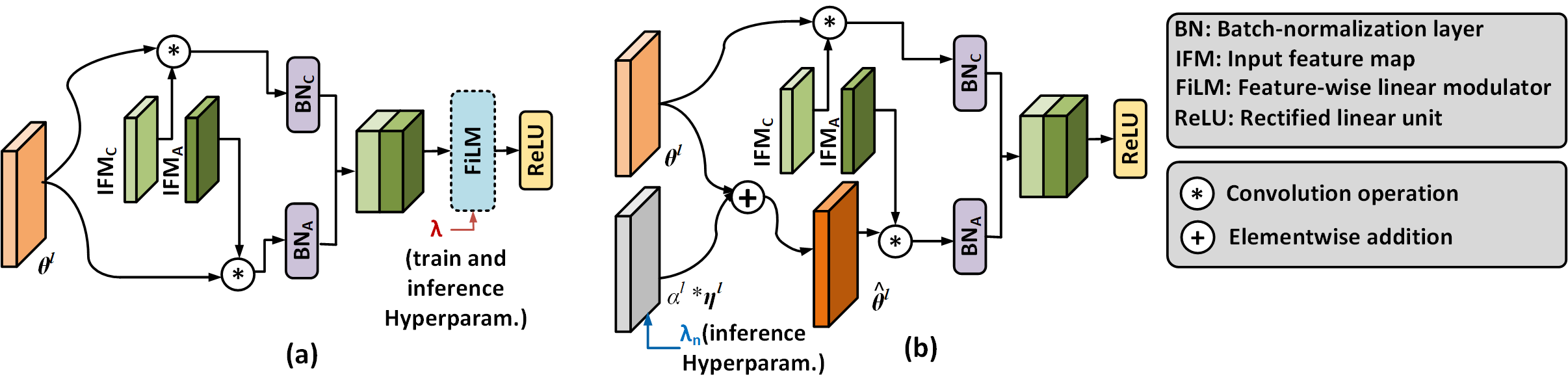}
  \centering
  \vspace{-4mm}
  \caption{Comparison of a conditional layer between (a) existing FiLM based approach in OAT and (b) proposed approach in FLOAT.}
  \label{fig:float_vs_oat}
  \vspace{-2mm}
\end{figure}

\subsection{Conditional Learning}
Conditional learning involves training a model with multiple computational paths that can be selectively 
enabled during inference \cite{wang2018skipnet}.
For example, \cite{teerapittayanon2016branchynet}, \cite{huang2017multi}, \cite{kaya2019shallow} enhanced a DNN model with multiple early exit branches at different architectural depths to allow early predictions of various inputs.
\cite{yu2018slimmable} introduced switchable BNs that enable the network to adjust the channel widths dynamically, providing an in-situ efficient trade-off between complexity and accuracy.
Recently, \cite{bulat2021bit} used switchable BNs 
to support runtime bit-width selection of a mixed-precision network.
Another conditional learning approach used feature transformation to modulate intermediate DNN features \cite{huang2017arbitrary}, \cite{yang2019controllable}, \cite{de2017modulating}, \cite{wang2020once}. 
In particular, \cite{wang2020once} used FiLM \cite{perez2018film} to adaptively perform a channel-wise affine transformation after each
BN stage that is controlled by the hyperparameter $\lambda$ of Equation \ref{eq:adv_loss}. Such conditional training that is able to yield models that can provide SOTA CA-RA trade-off on various $\lambda$ choices during inference are popularly known as Once-for-all adversarial training (OAT) \cite{wang2020once}. \begin{wrapfigure}{r}{0.30\textwidth}
\vspace{-2mm}
  \begin{center}
    \includegraphics[width=0.30\textwidth]{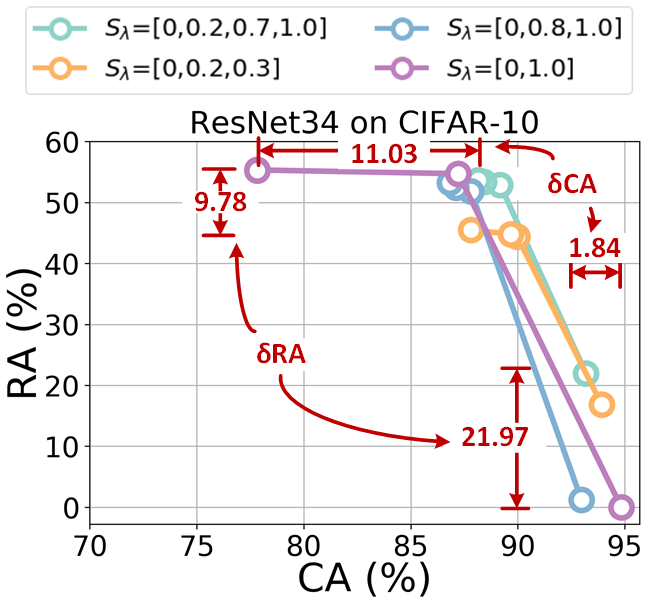}
  \end{center}
  \vspace{-5mm}
  \caption{Impact of various training $\lambda$ choices on the conditionally trained OAT.  During testing we use $S_{\lambda} = [0, 0.2, 0.7, 1.0]$.}
  \vspace{-4mm}
  \label{fig:lambda_impact}
\end{wrapfigure}

\textbf{Limitations of FiLM-based model conditioning.} Each FiLM module in OAT is composed of two fully-connected (FC) layers with leaky ReLU activation functions and dimensions that are integer multiples of the output feature-map channel size. Despite requiring a relatively small number of additional FLOPs, the FiLM module can significantly increase the number of model parameters and associated memory access cost \cite{horowitz20141}. Moreover, the increased number of layers can significantly increase training time and inference latency \cite{singh2019hetconv}, thus potentially prohibiting its use in real-time applications. 

Additionally, we investigated OAT's performance on the choice of the training $\lambda$ set ($S_{\lambda}$), as shown in Fig. \ref{fig:lambda_impact}. Interestingly, the CA and RA can vary up to $11.03\%$ and $21.97\%$, respectively. 
This implies that,\textit{OAT's performance may vary significantly based on both the size and specific values in $S_{\lambda}$.} In particular, the choice of $S_{\lambda}$ can significantly impact the robustness at $\lambda=0$, sometimes leading to no robustness.
%
%, OAT models show no robustness at $\lambda=0$, for some choices of  $S_{\lambda}$. 
This implies that to obtain models that yield near optimal CA-RA trade-offs, $S_{\lambda}$ must be carefully chosen, implying the need for an additional compute-heavy hyperparameter search or prior user expertise.  

\section{Proposed Approach}
\label{sec:approach}    
\subsection{FLOAT}

This section details our FLOAT training strategy. We refer to the conditions for a model being trained on either clean or adversarial images as the two training \textit{boundary conditions}. During training, we use a binary conditioning parameter $\lambda$ to force the model to focus on either of these two conditions, \textit{removing the need to search a more fine-grained set of $\lambda$ choices}. 
%We first perform conditional learning for a model, that can allow an user to focus on either of these two boundary conditions, conditioned using the hyperparameter $\lambda$ able take only binary values. 
\begin{comment}
\begin{wrapfigure}{r}{0.36\textwidth}
\vspace{-6mm}
  \begin{center}
    \includegraphics[width=0.35\textwidth]{Figs/scatter_distribution.jpg}
  \end{center}
  \vspace{-5mm}
  \caption{Comparison of the normalization statistics for a ResNet34 trained on CIFAR-10 with two different $\lambda$s. We plot the distribution by taking the running mean and running variance statistics from the last BN placed before the FC.}
  \vspace{-4mm}
  \label{fig:normal_dist}
\end{wrapfigure}
\end{comment}

%
To formalize our approach, consider a $L$-layer DNN parameterized by $\bm{\Theta}$ and let $\bm{\theta}^l \in \mathbb{R}^{k^l \times k^l \times C^l_i \times C^l_o}$ represent the layer $l$ weight tensor, where $C^l_o$ and $C^l_i$ represent the number of filters and channels per filter, respectively, and $k^l$ represents the kernel height/width.
 %of the weight tensor. 
We transform each parameter of $\bm{\theta}^l$, by adding a noise tensor $\bm{\eta}^l \in \mathbb{R}^{k^l \times k^l \times C^l_i \times C^l_o}$ scaled by a parameter $\alpha^l$
and conditioned by $\lambda$, as follows,
 %a transformed weight tensor is given by   
 \vspace{-4mm}
\begin{align}
    \hat{\theta}^l = \theta^l + \lambda\cdot\alpha^l\cdot\eta^l; \; \; \; \eta^l \mathord{\sim} \mathcal{N}(0, (\sigma^l)^2).
\end{align}
Note that the standard deviation $\sigma^l$ of the noise matches that of its weight tensor.
%set to the stan$\theta^l$
%is the standard deviation of the normally distributed noise. 
$\lambda = 0$ and $1$ generate the original weight tensor and its noisy variant, respectively. 

As illustrated in Algorithm \ref{alg:floats}, we train our models by partitioning an image batch $\mathcal{B}$ into two equal sub-batches $\mathcal{B}_1$ and $\mathcal{B}_2$, one with clean ($IFM_C$) images and the other with perturbed variants ($IFM_A$) (lines 5 and 7 in Algorithm \ref{alg:floats}). We use the PGD-7 attack to generate perturbations on the image batch $\mathcal{B}_2$.
As illustrated in Fig. \ref{fig:float_vs_oat}(b), the original and noisy weight tensors are convolved only with clean and perturbed variants, respectively. Note that the noise scaling factor $\alpha^l$ (line 10) is trainable and its magnitude can be different in each layer to minimize the total training loss. 
%Because we only add one scalar per layer, we incur negligible increase in model parameters. 
%
The post-convolution feature maps for clean and adversarial inputs can differ significantly in their respective mean and variances \cite{xie2020adversarial}, \cite{xie2019intriguing}. Therefore, the use of a single BN to learn both distributions may limit the model's performance \cite{wang2020once}. 
To solve this problem, we extend the $\lambda$-conditioning to choose between two BNs, $BN_C$ and $BN_A$, dedicated for $IFM_C$ and $IFM_A$, respectively. 

%Algorithm \ref{alg:float} details proposed training strategy.
\begin{wrapfigure}{r}{0.34\textwidth}
\vspace{-2mm}
  \begin{center}
    \includegraphics[width=0.34\textwidth]{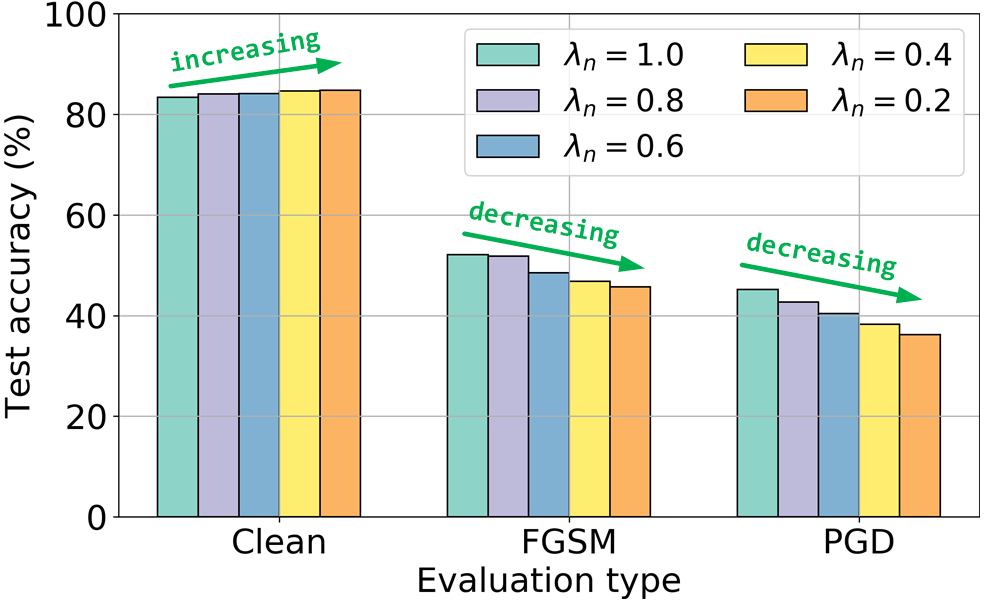}
  \end{center}
  \vspace{-5mm}
  \caption{Post-training model performance on both clean and gradient-based attack-generated adversarial images, with different noise  re-scaling factor $\lambda_n$.}
  \vspace{-3mm}
  \label{fig:motivational_lambda_rescaling}
\end{wrapfigure}
Our approach differs from previous efforts in several ways. Earlier research 
performed noise-injection via regularization \cite{bietti2018regularization}, \cite{lecuyer2019certified} and perturbed weight tensors \cite{he2019parametric} to boost model robustness at the cost of  a significant accuracy drop on clean images. In contrast, we use noise tensors to transform a shared weight tensor and yield a model that can be configured in-situ to provide SOTA accuracy on either clean or perturbed images.
Our approach is similar to $\lambda$-conditioning used by \cite{wang2020once}. However, instead of transforming activations
using added FiLM-based layers trained with multiple values of $\lambda$ \cite{wang2020once}, we transform weight tensors using added noise conditioned by
binary $\lambda$. Compared to
\cite{wang2020once}, we thus require models 
with significantly fewer parameters and training scenarios, yielding faster training (up to $1.43\times$).

\begin{comment}
\begin{algorithm}[t]
\small
\SetAlgoLined
\DontPrintSemicolon
\KwData{Training set $\bm{X} \mathord{\sim} D$, model parameters $\bm{\Theta}$, trainable noise scaling factor $\bm{\alpha}$, binary conditioning parameter $\lambda$, mini-batch size $\mathcal{B}$.}
\textbf{Output:} trained model parameters $\bm{\Theta}$, $\bm{\alpha}$.\\
\For{$\text{i} \leftarrow 0$ \KwTo \KwTo {$ep$}}
{
    \For{$\text{j} \leftarrow 0$ \KwTo {${n_{\mathcal{B}}}$}}
    {
        $\text{Sample clean image-batch of size } \mathcal{B}/2$ ($\bm{X}_{0:{\mathcal{B}/2}}$, $\bm{Y}_{0:{\mathcal{B}/2}}$) $\mathord{\sim} D$\;
        $\mathcal{L}_C \leftarrow \text{computeLoss}({\bm{X}_{0:{\mathcal{B}/2}}}, {\bm{\Theta}}, {\lambda}=0, \bm{\alpha}; \bm{Y}_{0:{\mathcal{B}/2}}) \text{ // condition to use weights w/o noise}$\;
        $\hat{\bm{X}}_{{\mathcal{B}/2}:{\mathcal{B}}} \leftarrow \text{createAdv}(\bm{X}_{{\mathcal{B}/2}:{\mathcal{B}}}, \bm{Y}_{{\mathcal{B}/2}:{\mathcal{B}}}) \text{ // adversarial image creation}$\;
        $\mathcal{L}_A \leftarrow \text{computeLoss}({\hat{\bm{X}}_{{{\mathcal{B}/2}}: {\mathcal{B}}}}, {\bm{\Theta}}, {\lambda}=1, \bm{\alpha}; \bm{Y}_{{{\mathcal{B}/2}}: {\mathcal{B}}}) \text{ // condition to use transformed weights}$\;
        
        $\mathcal{L} \leftarrow 0.5*\mathcal{L}_C + 0.5*\mathcal{L}_A$\;
        $\text{updateparam}(\bm{\Theta}, \bm{\alpha},\nabla_{\mathcal{L}})$\;
    }
}
 \caption{FLOAT Algorithm}
 \label{alg:float}
\end{algorithm}
\end{comment}

\textbf{FLOAT generalization with noise re-scaling.}
One limitation of the FLOAT as proposed above is that it allows the user to choose between two boundary conditions only. This limits applicability when the user is not confident about which condition to use during inference. To motivate more continuous in-situ conditioning, we analyze a ResNet20 model with noisy weight tensors trained with PGD-AT on CIFAR-10 \cite{he2019parametric}. Post-training, we re-scaled $\alpha^l$ for each layer $l$, using a new floating-point parameter $\lambda_n$ to yield $\lambda_n \cdot \alpha^l$.
%
%using a re-scaling factor $\lambda_n$. For example, for $\lambda_n = 1.0$ we used $\alpha^l$ for layer $l$, and similarly for $\lambda_n = 0.4$, we used $0.4*\alpha^l$ for the same layer. 
Interestingly, as shown in Fig. \ref{fig:motivational_lambda_rescaling}, as the re-scaling factor decreases, the model robustness decreases and the clean-image accuracy increases.

\textcolor{black}{Based on this observation, we introduce a practical means of post-training in-situ calibration by adding a re-scaling parameter $\lambda_n$ to the inference model}\footnote{Note that $\lambda_n$ is a continuous variable between 0 and 1 where as $\lambda$ is binary. $\lambda_n=0$ and $\lambda_n=1$ matches the training boundary conditions. OAT, on the other hand, uses a single variable $\lambda$ that
can be any floating point value in $[0,1]$ during both training and inference.}.
%as it can assume any floating point value in $[0,1]$.}. 
\textcolor{black}{This allows us to enable a practical accuracy-robustness trade-off in FLOAT during inference.}
We also define a threshold $\lambda_{th}$ such that for $\lambda_n >\lambda_{th}$ we select $BN_A$ to perform inference and select $BN_C$ otherwise. \cite{wang2020once} selected $BN_C$ and $BN_A$ when $\lambda = 0$ and $\lambda > 0$, respectively. We follow a similar approach by setting $\lambda_{th} = 0$.
%ECCV 2022 newly added
\subsection{FLOAT Extension to Model Compression via Pruning}
Pruning is a particular form of model compression that has been effective in reducing model size and compute complexity for large DNNs for resource-constrained deployment \cite{chen2021chasing,kundu2021attentionlite,liu2018rethinking,he2018amc}. Motivated by these results, we incorporate a form of pruning called sparse learning\footnote{Every update of the model happens sparsely, meaning only a fraction of the weights are updated, while other remains as zero.} \cite{kundu2021attentionlite} into FLOAT, which we refer to FLOAT sparse-$irregular$ (FLOATS-$i$). The resulting approach not only provides a CA-RA trade-off, but also meets a target global parameter density $d$. 
In particular, FLOATS ranks every layer based on the normalized momentum of its non-zero parameters. Based on this ranking, FLOATS dynamically allocates more weights to layer that have larger momentum and fewer weights to other layers, while maintaining the global density constraint. To be more precise, let the binary pruning mask be parameterized by the set $\mathbf{\Pi}$ with elements $\bm{\pi}^l$ representing the mask tensor for layer $l$. The fraction of 1s in $\bm{\pi}^l$ is proportional to its relative layer importance evaluated through momentum. During training, the total cardinality of the masked parameters \textit{always} satisfies the following constraint 
\vspace{-2mm}
\begin{align}
\sum_{l=1}^L \texttt{card}({\bm{\theta}^{l}} \odot \bm{\pi}^l) \le d \sum_{l=1}^L \texttt{card}(\bm{\theta}^l).
\end{align}
\vspace{-2mm}
%to have potential speed up on the target hardware \cite{liu2018rethinking}.
\begin{algorithm}[t]
\scriptsize
\SetAlgoLined
\DontPrintSemicolon
\KwData{Training set $\bm{X} \mathord{\sim} D$, model parameters $\bm{\Theta}$, trainable noise scaling factor $\bm{\alpha}$, binary conditioning parameter $\lambda$, mini-batch size $\mathcal{B}$, global parameter density $d$, initial mask $\mathbf{\Pi}$, prune type (irregular/channel) $t_p$.},
\textbf{Output:} trained model parameters with density $d$.\\
$ \bm{\Theta} \leftarrow \texttt{applyMask(}\bm{\Theta},{\mathbf{\Pi}} \text{)} $\;

\For{$\text{i} \leftarrow 0$ \KwTo \KwTo {$ep$}}
{
   
    \For{$\text{j} \leftarrow 0$ \KwTo {${n_{\mathcal{B}}}$}}
    {
     $ \mathcal{B}/2$ ($\bm{X}_{0:{\mathcal{B}/2}}$, $\bm{Y}_{0:{\mathcal{B}/2}}$) $\mathord{\sim} D$\;
        $\mathcal{L}_C \leftarrow \texttt{computeLoss}({\bm{X}_{0:{\mathcal{B}/2}}}, {\bm{\Theta}}, {\lambda}=0, \bm{\alpha}; \bm{Y}_{0:{\mathcal{B}/2}}) $\;
         $\hat{\bm{X}}_{{\mathcal{B}/2}:{\mathcal{B}}} \leftarrow \texttt{createAdv}(\bm{X}_{{\mathcal{B}/2}:{\mathcal{B}}}, \bm{Y}_{{\mathcal{B}/2}:{\mathcal{B}}}) $\;
        $\mathcal{L}_A \leftarrow \texttt{computeLoss}({\hat{\bm{X}}_{{{\mathcal{B}/2}}: {\mathcal{B}}}}, {\bm{\Theta}}, {\lambda}=1, \bm{\alpha}; \bm{Y}_{{{\mathcal{B}/2}}: {\mathcal{B}}})$\;
        
        $\mathcal{L} \leftarrow 0.5*\mathcal{L}_C + 0.5*\mathcal{L}_A$\;
        $\texttt{updateParam}(\bm{\Theta}, \bm{\alpha},\nabla_{\mathcal{L}},{\mathbf{\Pi}})$\;
    }
    {
        $\texttt{updateLayerMomentum(}\bm{\mu})$\;
        $ \texttt{pruneRegrow(}\bm{\Theta}, {\mathbf{\Pi}}, \bm{\mu}, d)\text{// Prune fixed \% of active and regrow fraction of}$\;
        $\text{inactive weights}$\;
        ${\mathbf{\Pi}} \leftarrow \texttt{updateMask(}{\mathbf{\Pi}}, t_p, \bm{\mu}\text{)}$\;
  }
}
\vspace{-1mm}
 \caption{FLOATS Algorithm}
 \label{alg:floats}
\end{algorithm}
\begin{figure}[!b]
    \includegraphics[width=0.90\textwidth]{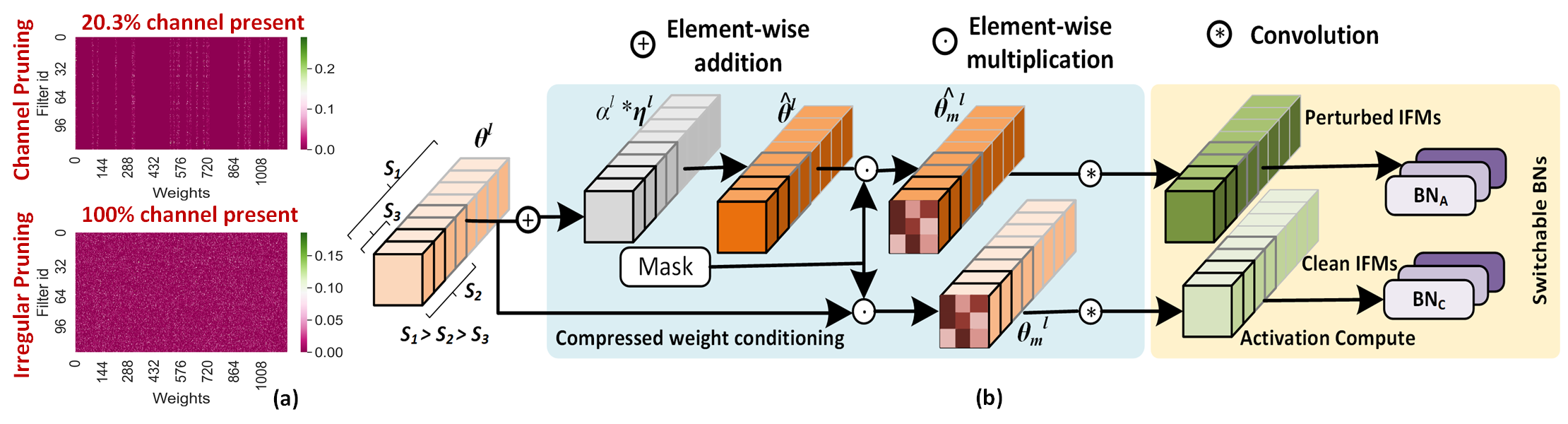}
  \centering
  \vspace{-4mm}
  \caption{\textcolor{black}{(a) Comparison of channel density (weights plotted in abs. magnitude) for FLOATS irregular and channel, for the $29^{th}$ CONV layer of WRN40-2 on STL10 while both are trained for $d = 0.3$. (b) Convolutional layer operation path for FLOATS slim. Note, the switchable BNs correspond to BNs for each SF.} }
  \label{fig:floats_i_vs_c_density_plot}
  \vspace{-4mm}
\end{figure}

To further ensure that the pruned models have structure and enable speed-up on a wide range of existing hardware \cite{liu2018rethinking}, we propose FLOATS-$c$ that performs $channel$ pruning. In FLOATS-$c$, for a layer $l$, we convert the 4D ${\bm{\theta}^{l}}$
to a 2D weight matrix with $C^l_o$ rows and $(k^l)^2C^l_i$ columns that is further partitioned in to $C^l_i$ sub-matrices of $C^l_o$ rows and $(k^l)^2$ columns. To evaluate the channel importance, we compute the Frobenius norm (F-norm) of each sub-matrix $c$ by computing ${f}^l_c$ = $||{\bm{\theta}^{l}_{:,c,:,:}}||^{2}_F$. We then keep or remove a channel based on the ranking of $f^l_c$'s, enabling pruning at the channel level.
%, that essentially is a stricter constraint than the irregular counterpart, and can provide speed-up. In particular, 
As depicted in Fig. \ref{fig:floats_i_vs_c_density_plot}(a), the weight heatmaps show that for the same layer FLOATS-$c$ can yield only $20.3\%$ non-zero channels, while FLOATS-$i$ retains all the channels. In fact for the same target $d$, the channel density can be $10\times$ lower for some layers as compared to that in FLOATS-$i$. We note that this large scale channel reduction sometimes comes at a non-negligible  accuracy drop as shown in Table \ref{tab:floats_perfromance_res34}. 

\textbf{A globally structured locally irregular pruning.} To simultaneously benefit from  aggressive parameter reduction via irregular pruning and width reduction via channel pruning, while maintaining high accuracy, we propose a form of hybrid compression called FLOATS slim. FLOATS slim leverages the idea of slimmable networks \cite{yu2018slimmable} to train a model with channel widths that are scaled by a global channel slimming-factor (SF). On top of this, we use FLOATS-$i$ to yield a locally irregular model with even fewer parameters for a specific SF. We perform both of these optimizations simultaneously, training with multiple SFs, including SF $=1$ (Algorithm detailed in the supplementary material). Note, unlike FLOATS-$c$, where different layers might have different SFs, FLOATS slim yields uniform SFs for all layers. However, in FLOATS slim, a model with SF$<1.0$ is trained as a shared-weight sub-network of the model with SF $=1.0$, contrasting FLOATS-$c$, where only one model of a specific $d$ is trained. Fig. \ref{fig:floats_i_vs_c_density_plot}(b) depicts the weight conditioned convolution operation in FLOATS slim.

\section{Experimental Results and Analysis}
\label{sec:expt}

\subsection{Experimental Setup}
\label{subsec:setup}
\textbf{Models and datasets.} 
To evaluate the efficacy of the presented algorithms, we performed detailed experiments on \textcolor{black}{five popular datasets, CIFAR-10, CIFAR-100 \cite{krizhevsky2009learning}, Tiny-ImageNet \cite{hansen2015tiny} with ResNet34 \cite{he2016deep}, SVHN \cite{netzer2011reading} with WRN16-8 \cite{zagoruyko2016wide}, and STL10 \cite{coates2011analysis} with WRN40-2 \cite{zagoruyko2016wide}}. 

\textbf{Hyperparameters and training settings.} In order to facilitate a fair comparison, for CIFAR-10, SVHN, and STL10 we used similar hyperparameter settings as \cite{wang2020once}\footnote{We followed the official repository https://github.com/VITA-Group/Once-for-All-Adversarial-Training}. \textcolor{black}{For CIFAR-100, we followed same hyperparameter settings as that with CIFAR-10. For Tiny-ImageNet we trained the model for 120 epochs with an initial learning rate of 0.1 an used cosine decay.} For adversarial image generation during training, we used the PGD-$k$ attack with $\epsilon$ and $k$ set to 8/255 and 7, respectively. We initialized the noise scaling-factor $\alpha^l$ for layer $l$ to $0.25$ as described in \cite{he2019parametric}. 
We used the PyTorch API \cite{paszke2017automatic} to implement our models and trained them on a Nvidia GTX Titan XP GPU.

\textbf{Evaluation metrics.} 
Clean (standard) accuracy (CA): classification accuracy on the original clean test images. %CA denotes the (default) accuracy. 
Robust Accuracy (RA): classification accuracy on adversarially perturbed images generated from the original test set. We use RA as the measure of robustness of a model. \textcolor{black}{To directly measure the robustness vs accuracy trade-off, we evaluated the clean and robust accuracy values of models generated through FLOAT at various $\lambda$ values and compared with those yielded through OAT and PGD-AT}. We used the average of the best CA and RA values over three different runs with varying random seeds, for each $\lambda$ value to report in our results.

\subsection{Performance of FLOAT}

\textbf{Sampling $\lambda_n$.} Unless stated otherwise, to evaluate the performance of FLOAT during validation we chose a set of $\lambda_n$s as $S_{\lambda_n} = \{0.0, 0.2, 0.7, 1.0\}$. Note that setting $\lambda_n$ to $0.0$ or $1.0$ corresponds to the values of $\lambda$ used during training. Also, we measure the accuracy of FLOAT using two different settings of $\lambda_{th}$, $0.0$ (similar to OAT) and $0.5$. For $\lambda_{th}=0.5$, we update the noise scaling factor by using the following simple equation 
\begin{align}
    \alpha^l_{new} = 
    \begin{cases}
        \alpha^l \cdot 2\cdot\lambda_n ; \text{ if }\lambda_n \leq 0.5\\
        \alpha^l \cdot 2\cdot(\lambda_n - 0.5) ; \text{ if } 0.5<\lambda_n \leq 1.0
    \end{cases}
    \label{eq:nose_scaling_lambda_5}
\end{align}
\begin{figure}[!t]
    \includegraphics[width=1.0\textwidth]{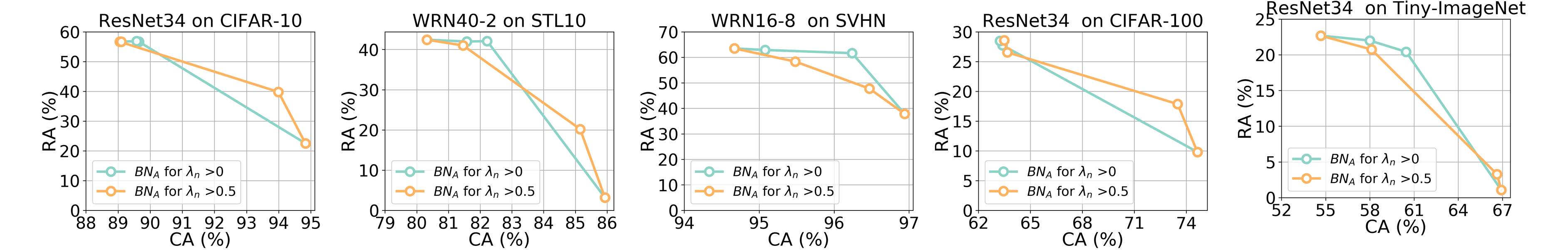}
  \centering
  \vspace{-3mm}
  \caption{\textcolor{black}{Performance of FLOAT on (a) CIFAR-10, (b) STL10, (c) SVHN, (d) CIFAR-100, and (e) Tiny-ImageNet with various $\lambda_n$ values sampled from $S_{\lambda_n}$ for two different $\lambda_{th}$ for $BN_C$ to $BN_A$ switching. The numbers in the bracket corresponds to (CA, RA) for the boundary conditions of $\lambda=0$ and $\lambda=1$. $\lambda_n$ varies from largest to smallest value from top-left to bottom-right point}.}
  \label{fig:float_bn_abl}
  \vspace{-4mm}
\end{figure}
% As depicted in Fig. \ref{fig:float_bn_abl} (a)-(d), the FLOAT models generalize well to yield a continuous accuracy-robustness trade-off. Also, for all the datasets, $\lambda_{th} = 0.5$ yields a more gradual transition between the two boundary conditions. In particular, compared to $\lambda_{th}=0.0$, this yields a CA improvement of $4.65\%$ with reduction in RA of $11.18\%$ averaged over four datasets (for $\lambda_n = 0.2$). The improved clean accuracy here can be attributed to the use of $BN_C$. However, in comparison to CA-RA performance for $\lambda_n = 0.0$, we see a drop in CA and improvement in RA when $\lambda_n=0.2$ that can be attributed to the use of noisy weights (refer to Eq. \ref{eq:nose_scaling_lambda_5}), during inference. From this we can conclude that, an user caring more about clean image performance, should set $\lambda_{th} > 0.0$ to see less abrupt drop in CA. 

As depicted in Fig. \ref{fig:float_bn_abl} \textcolor{black}{(a)-(e)}, the FLOAT models generalize well to yield a semi-continuous accuracy-robustness trade-off. Also, across all the datasets, $\lambda_{th} = 0.5$ yields a more gradual transition between the two boundary conditions. Consider the setting where $\lambda_n = 0.2$. With $\lambda_{th} = 0.5$, we observe a \textcolor{black}{$4.95\%$} improvement in CA and a reduction in RA of \textcolor{black}{$12.37\%$} on average over all \textcolor{black}{five} datasets when compared with $\lambda_{th} = 0.0$. The improvement in clean accuracy here can be attributed to the use of $BN_C$. However, this configuration shows a drop in CA and an improvement in RA when compared to the configuration where $\lambda_n = 0.0$. This can be attributed to the use of noisy weights (refer to Eq. \ref{eq:nose_scaling_lambda_5}) during inference. Thus, it can be concluded that a user who cares more about clean image performance than adversarial robustness should set $\lambda_{th} > 0.0$ to see a less abrupt drop in CA. \textcolor{black}{Note that, because the generation of adversarial images is noisy, it is not always true that increasing $\lambda$ will always significantly improve robustness. Consequently, in some cases, we obtain improved clean image performance without a significant drop in robustness.}

\subsection{Comparison with OAT and PGD-AT}
We trained the benchmark models following OAT and PGD-AT with $\lambda$s sampled from a set $S_{\lambda} = S_{\lambda_n}$ on three datasets, CIFAR-10, SVHN, and STL10.

\begin{figure}[!t]
    \includegraphics[width=0.78\textwidth]{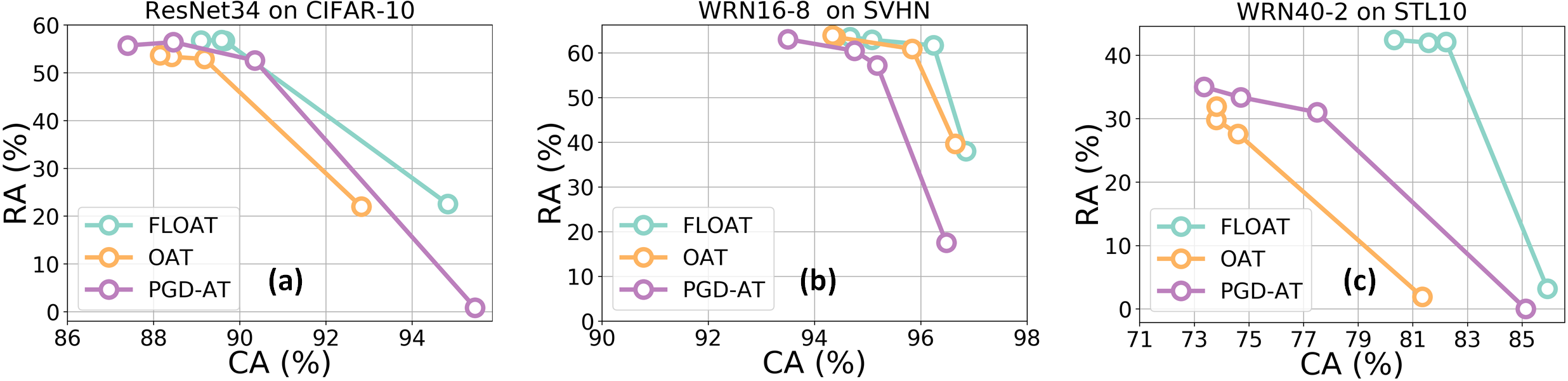}
  \centering
  \vspace{-4mm}
  \caption{Performance comparison of FLOAT with OAT and PGD-AT generated models on (a) CIFAR10, (b) SVHN, and (c) STL10. $\lambda$ varies from largest to smallest value in $S_{\lambda}$ for the points from top-left to bottom-right.}
  \label{fig:float_vs_oat_vs_pgd_com}
  \vspace{-4mm}
\end{figure}

\textbf{Discussion on CA-RA trade-off.}  
Fig. \ref{fig:float_vs_oat_vs_pgd_com}(a)-(c) show the comparison of FLOAT with OAT and PGD-AT in terms of CA-RA trade-offs. The FLOAT models show similar or superior performance at the boundary conditions as well as at intermediate sampled values of $\lambda$. In particular, compared to OAT and PGD-AT models, FLOAT models can provide an improved RA of up to $14.5\%$ (STL10, $\lambda=0.2$) and $34.92\%$ (CIFAR-10, $\lambda=0.0$), respectively. FLOAT also provides improved CA of up to $6.5\%$ (STL10, $\lambda=1.0$) and $6.96\%$ (STL10, $\lambda=1.0$), compared to OAT and PGD-AT generated models, respectively. \textcolor{black}{Interestingly, for both FLOAT and OAT, in all the plots we generally see a sharp drop in robustness while moving from top-left to bottom-right. This can be attributed to to the switch from $BN_A$ to $BN_C$ based on the $\lambda_{th}$, in the forward pass of the inference model.}

\begin{figure}[!t]
    \includegraphics[width=0.82\textwidth]{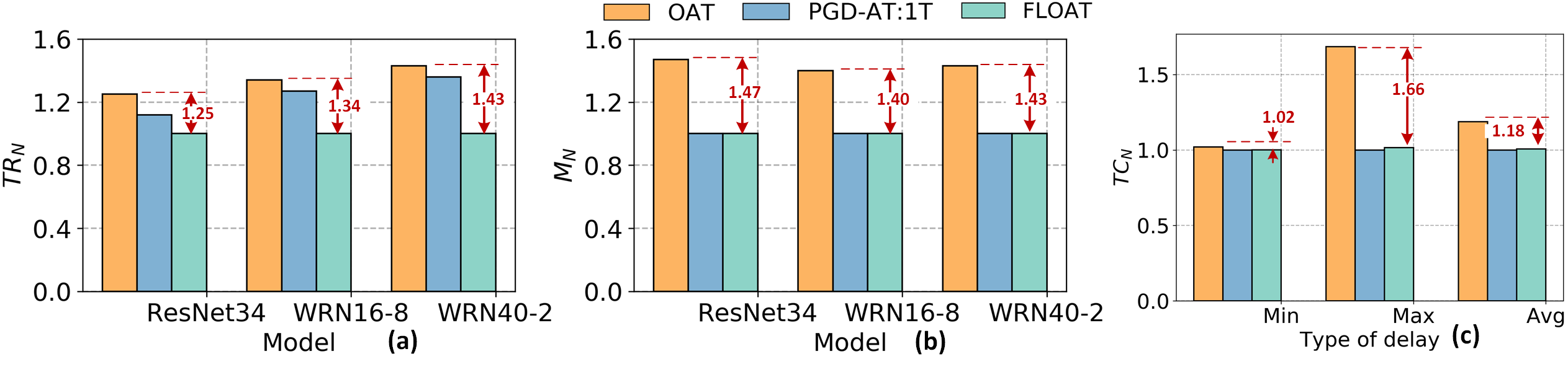}
  \centering
  \vspace{-4mm}
  \caption{\textcolor{black}{Comparison of FLOAT with OAT and PGD-AT in terms of (a) normalized training time per epoch and (b) model parameter storage  (neglecting the storage cost for the BN and $\bm\alpha$) (c) CONV layer compute delay on conventional ASIC (using the delay model of Eq. 7, 8, and 9) architecture \cite{ali2020imac} evaluated on ResNet34 for CIFAR-10. Note here, PGD-AT:1T yields 1 model for a specific $\lambda$ choice.} }
  \label{fig:float_oat_training_time_memory}
  \vspace{-6mm}
\end{figure}

\textbf{Discussion on  training time and inference latency.}  
Due to the presence of the additional FiLM modules, OAT requires more time than standard PGD-AT to train. However, a single PGD-AT training can only provide a fixed accuracy-robustness trade-off. For example, to have trade-off with 4 different $\lambda$s PGD-AT training time increases proportionally by a factor of $4$.
%, making this training strategy extremely inefficient for yielding multiple trade-off options. 
FLOAT, on the contrary, due to absence of additional layers, trains faster than OAT.
%, yet can yield in-situ trade-offs. 
In particular, Fig. \ref{fig:float_oat_training_time_memory}(a) shows the normalized per-epoch training time 
(averaged over $200$ epochs) of \textcolor{black}{OAT and PGD-AT are, respectively,
up to $1.43\times$ and $1.37\times$ slower than FLOAT}. 

Network latency increases with the increase in the number of layers for both standard and mobile GPUs \cite{li2021npas}, \cite{singh2019hetconv}, primarily because layers are operated on sequentially \cite{singh2019hetconv}. The additional FiLM modules in OAT significantly increase the layer count. For example, for each bottleneck layer in ResNet34, OAT requires two FiLM modules, yielding a total of four additional FCs per bottleneck.
%each bottleneck layer of ResNet34 incurs 2 additional FiLM module, thus four additional layers. 
On the other hand, FLOAT, \textcolor{black}{similar to a single PGD-AT trained model,} requires no additional layers or associated latency, making it more attractive for real-time applications.  

\textbf{Discussion on model parameter storage cost.}
Unlike OAT, where the  FiLM layer FCs significantly increase the parameter count, the additional BN layers and scaling factors of FLOAT represent a negligible increase in parameter count. In particular, assuming parameters are represented with 8-bits, a FLOAT ResNet34 has only 21.28 MB memory cost compared to 31.4MB for OAT. Fig. \ref{fig:float_oat_training_time_memory}(b) shows that FLOAT models, \textcolor{black}{similar to PGD-AT:1T,} can yield up to $1.47\times$ lower memory.

\textbf{Discussion on FLOPs.} Compared to the standard PGD-AT, FLOAT incurs additional compute cost of addition of noise with the weight tensor during forward pass. For example, for ResNet34 with $\mathord{\sim}21.28$ M parameters, FLOAT needs similar number of additions for noisy weight transformation. However, compared to the total operations of $\mathord{\sim}1.165$  GFLOPs, the transformation adds on $1.182\%$ additional computation. Moreover, as a single addition can be up to $32\times$ cheaper than a single FLOP \cite{horowitz20141}, we can gracefully ignore such transformation cost in terms of FLOPs. OAT, on the other hand, also incurs negligibly less FLOPs overhead of up to only $\mathord{\sim}1.7\%$  \cite{wang2020once}.

\textbf{Discussion on compute delay in Von-Neumann ASIC architecture.} In a conventional Von-Neumann architecture a neural network algorithm can be broken down into two dominant operations: memory read and Multiply-accumulate (MAC). Based on the assumption that these operations are sequential, as in \cite{ali2020imac,kang2018memory}, the layer delay to compute $C^l_o$ output-features can be estimated 
\vspace{-2mm}
\begin{align}
    \tau_{conv} \approx \lceil \frac{(k^l)^2C^l_iC^l_o}{(B_{IO}/B_W)N_{bank}} \rceil \tau_{read} + \lceil \frac{(k^l)^2C^l_iC^l_o}{N_{Mult}} \rceil H^l_oW^l_o\tau_{mult}.
\end{align}
where $B_{IO}$ is the memory input-output (IO) bandwidth and $B_{W}$ is the bit-width of each weight stored in memory. $N_{bank}$ and $N_{mult}$ corresponds to the number of hardware memory banks and multiply units \cite{kang2018memory}. A single memory read and multiply operation time is denoted by $\tau_{read}$ and $\tau_{mult}$, respectively. Their values for a 65nm CMOS process technology are $9ns$ and $4ns$, respectively \cite{kang2018memory}. Based on similar assumptions, the delay model for modified CONV layer $l$ for FLOAT ($\tau^{F}_{conv}$) and OAT ($\tau^{O}_{conv}$) can be estimated as
\vspace{-0mm}
\begin{align}
    \tau^{F}_{conv} \approx \lceil \frac{(k^l)^2C^l_iC^l_o}{(B_{IO}/B_W)N_{bank}} \rceil \tau_{read} + \lceil \frac{(k^l)^2C^l_iC^l_o}{N_{Mult}} \rceil (1 + H^l_oW^l_o)\tau_{mult},
    \label{eq:float_delay}
\end{align}
\vspace{-0mm}
\begin{align}
    \nonumber  \tau^{O}_{conv} \approx \lceil \frac{(k^l)^2C^l_iC^l_o + 2C^l_o + 4(C^l_o)^2}{(B_{IO}/B_W)N_{bank}} \rceil \tau_{read} + \lceil \frac{(k^l)^2C^l_iC^l_o}{N_{Mult}} \rceil H^l_oW^l_o\tau_{mult} + \\ 
    \lceil \frac{2C^l_o + 4(C^l_o)^2}{N_{Mult}} \rceil \tau_{mult}.
    \label{eq:oat_delay}
\end{align}
Here, the first term corresponds to the read delay and remaining term(s) correspond to the delay associated with the multiplications. We ignore the energy associated with reading $\alpha^l$ because it is negligible compared to the read energy for the other model parameters. 
Based on these Eqs, Fig. \ref{fig:float_oat_training_time_memory}(c) shows the minimum, maximum, and average normalized delays with respect to the $\tau_{conv}$. In particular, conditional CONV layer delay of FLOAT can be up to $1.66\times$ faster compared to that of OAT, illustrating its efficacy on conventional architecture. 

\subsection{Performance of FLOATS}
\vspace{-2mm}
\begin{table}[!h]
\begin{center}
\scriptsize\addtolength{\tabcolsep}{-1.5pt}
\begin{tabular}{c|c|c|c|c|c|c|c|c}
\hline
\multicolumn{1}{c|}{ Algorithm}  &\multicolumn{2}{c|}{ Acc. $\%$ ($\lambda=0.0$)} &\multicolumn{2}{c|}{ Acc. $\%$ ($\lambda=1.0$)} & \multicolumn{1}{c|}{ CR $\uparrow$} & \multicolumn{1}{c|}{ CRF $\uparrow$} & Reduced & Potential\\
 \cline{2-5}
 & \multicolumn{1}{p{1.2cm}|}{ CA} & \multicolumn{1}{p{1.2cm}|}{ RA} & \multicolumn{1}{p{1.2cm}|}{ CA} & \multicolumn{1}{p{1.2cm}|}{ RA} & \multicolumn{1}{p{1.2cm}|}{} & \multicolumn{1}{p{1.0cm}|}{} & \multicolumn{1}{p{1.2cm}|}{storage} & speed-up\\
 \hline
FLOAT         & \textbf{94.83} & \textbf{22.52} & \textbf{89.1} & \textbf{56.71} & $1\times$ & $1 \times$ & \xmark & \xmark\\
\hline
FLOATS-$i$             & 94.12 & 18.7 & 88.6 & 55.92 & $\textbf{10}\times$ & $1 \times$ & \cmark \cmark & \xmark\\
\hline
FLOATS-$c$             & 93.84 & 17.2 & 86.87 & 53.2 & $2.94\times$ & $1.54 \times$ & \cmark & \cmark\\
\hline
FLOATS slim             & 94.26 & 19.1 & 88.9 & 55.44 & $4.76\times$ & $\textbf{2} \times$ & \cmark \cmark & \cmark\\
\hline
\end{tabular}
\end{center}
\caption{Performance comparison between different compressed FLOAT variants trained on CIFAR-10 with ResNet34. \cmark\cmark, \cmark, \text{ and} \xmark  \text{ indicate} aggressive, non-aggressive, and no reduction, respectively, compared to the baseline of FLOAT.}
\vspace{-2mm}
\label{tab:floats_perfromance_res34}
\end{table}
Table \ref{tab:floats_perfromance_res34} shows the performance of FLOATS with irregular, channel, and slimmable compression. The FLOATS slim model was trained with two representative SFs of $1.0$ and $0.5$ with a global target density $d = 0.3$. We report its performance with SF$=0.5$. Here, \textit{compression ratio}(CR) and $\textit{channel reduction factor}$ (CRF) are computed as $\frac{1}{d}$ and $\frac{1}{\% \text{ of total channels present}}$, respectively. Compared to FLOAT-$c$, FLOATS slim requires $1.62\times$ less storage, results in up to $2\times$ speed-up, and yields $2.24\%$ higher classification accuracy. Moreover, FLOATS slim provides us with a unique three-way trade-off between robustness, accuracy, and complexity, requiring only single training pass. 

\begin{figure}[!t]
    \includegraphics[width=0.95\textwidth]{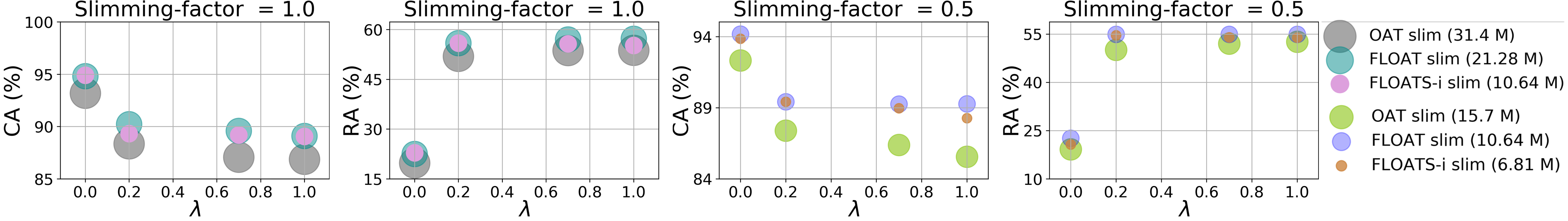}
  \centering
  \vspace{-2mm}
  \caption{Performance comparison of FLOAT slim, FLOATS-$i$ slim with OAT slim. We used ResNet34 on CIFAR-10 to evaluate the performance.}
  \label{fig:floats_vs_oats_c10}
  \vspace{-4mm}
\end{figure}

Fig. \ref{fig:floats_vs_oats_c10} illustrates the efficacy of FLOAT slim compared to OAT slim. 
%Because OAT does not support sparsity, we include FLOAT slim results without any sparsity. 
FLOAT slim provides significantly improved performance for all tested values of $\lambda$ for both the SFs. In particular, FLOAT slim yields up to $3.6\%$ higher accuracy. Adding sparsity, FLOATS slim yields similar accuracy improvement with up to $2.95\times$ less parameters. Moreover, GPU hardware measurements show that our slimmable training is $1.90\times$ faster training OAT slim. 
\subsection{Generalization on Various Perturbation Techniques}
\begin{figure}[!t]
    \includegraphics[width=0.75\textwidth]{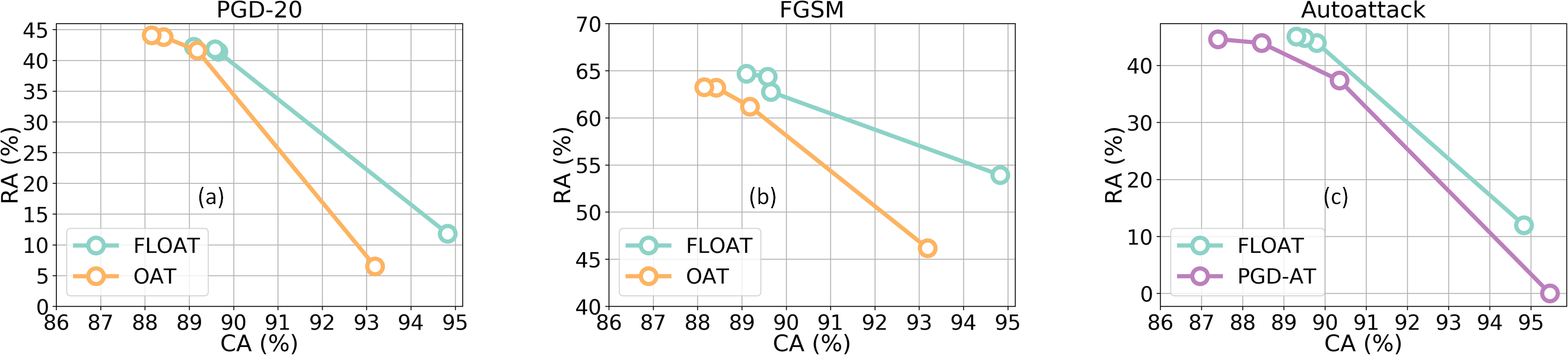}
  \centering
  \vspace{-0mm}
  \caption{Performance comparison of FLOAT with OAT on (a) PGD20 and (b) FGSM attack generated images. (c) CA-RA plot of FLOAT vs. PGD-AT on autoattack. All evaluations are done with ResNet34 on CIFAR-10. $\lambda$ varies from largest to smallest value in $S_{\lambda}$ for the points from top-left to bottom-right.}
  \label{fig:float_vs_oat_vs_pgdat_pgd20_fgsm_auto}
  \vspace{-4mm}
\end{figure}

To demonstrate the generalization of FLOAT models on different attacks, we show their performance on images adversarially-perturbed through PGD-20 and FGSM attacks. We follow \cite{wang2020once} to generate the PGD-20 perturbations and set the number of steps to 20, keeping other hyperparameters the same as PGD-7. For FGSM, we make $\epsilon = 8/255$ following \cite{wang2020once}. As shown in Fig. \ref{fig:float_vs_oat_vs_pgdat_pgd20_fgsm_auto}(a)-(b), under both the attacks, FLOAT can achieve in-situ accuracy-robustness trade-offs similar to that of OAT. \textcolor{black}{Moreover, we have analyzed FLOAT's robustness with an ensemble of parameter-free attacks, namely the `random' variant of autoattack \cite{croce2020reliable}.\footnote{We have followed the official repo https://github.com/fra31/auto-attack to generate the attack.}. Details of the autoattack hyperparameters are provided in the Supplementary Materials. As depicted in \ref{fig:float_vs_oat_vs_pgdat_pgd20_fgsm_auto}(c), compared to the PGD-AT yielded models, FLOAT consistently yields better RA with similar or improved CA.}
\section{Conclusions}
\label{sec:conc}
This paper addresses the largely unexplored problem of enabling an in-situ inference trade-off between accuracy, robustness, and complexity. We propose a fast learnable once-for-all adversarial training (FLOAT) which uses model conditioning to capture the different feature-map distributions corresponding to clean and adversarial images. FLOAT transforms its weights using conditionally added scaled noise and dual batch normalization structures to distinguish between clean and adversarial images. The approach avoids increasing the layer count, unlike other state of the art alternatives, and thus does not suffer from increased network latency. %we neither increased any layer latency nor did we suffer from any significant increase in model parameters. 
We then extend FLOAT to include sparsity to further reduced complexity and latency providing an in-situ trade-off including model complexity.
%in-situ trade-off between accuracy, robustness, and model complexity. 
\textcolor{black}{Extensive experiments show FLOAT’s superiority in terms of improved CA-RA performance, reduced parameter count, and faster training time.}
\section{Broader Impact}
\label{sec:broad}
DNNs are well-known to be susceptible to adversarial images \cite{szegedy2013intriguing}. As their application grows in various safety-critical applications, including autonomous driving \cite{bojarski2016end}, medical imaging \cite{han2021advancing}, and household robotics \cite{tritschler_2021}, achieving model robustness without sacrificing clean image accuracy is increasingly important. This is particularly important when the image scenario is dynamic. 
%This is particularly important for scenarios that can change frequently such as in the case of a household robots and military drones.
%
%against adversarial images along with the retention of performance on clean images, is on the rise. 
Moreover, the increasingly portable nature of these AI-enabled devices introduces stringent storage and energy-budget limitations. To address these challenging problems, this paper presents FLOAT models that can be configured in-situ to dynamically adjust the model's accuracy, robustness, and complexity based on the image scenario. 
%trade-off and can train up to $1.52\times$ faster with up to $1.47\times$ fewer parameters compared to the existing alternatives. 
Our approach does not require iterative (re-)training as with standard PGD-AT and trains significantly faster than SOTA alternatives. Moreover, for inference-only devices, our approach circumvents the need for re-loading alternate model parameters from the cloud to support different scenarios, avoiding the potentially significant data-transfer costs. 
With the increase of high-stake real-world applications that require robustness, we believe this research will form the foundation for practical AI-driven applications that can efficiently adapt to their environment. \textcolor{black}{Finally, we hope this paper will motivate further work aimed at enhancing the continuity of the accuracy-robustness trade-off and developing a theoretical basis that can explain the benefits and limitations of FLOAT and conditional learning in general.}
%to support dynamic performance switch based on various surrounding requirements.
%No ack mention during review
%\subsubsection*{Acknowledgments}
%This work was supported in parts by Intel Labs, USA.
\begin{comment}
\section{Reproducibility}
\label{sec:reproduce}
We have detailed our training algorithms for both FLOAT and FLOATS in Algorithm \ref{alg:float} and \ref{alg:floats}, respectively. We have further provided detailed hyperparameter and model information in Section \ref{subsec:setup} as well as in the Appendix \ref{subsec:hyper}. Finally, the code for this project will be made publicly available shortly.
\end{comment}

\bibliography{iclr2022_conference}

\begin{thebibliography}{47}
\providecommand{\natexlab}[1]{#1}
\providecommand{\url}[1]{\texttt{#1}}
\expandafter\ifx\csname urlstyle\endcsname\relax
  \providecommand{\doi}[1]{doi: #1}\else
  \providecommand{\doi}{doi: \begingroup \urlstyle{rm}\Url}\fi

\bibitem[Ali et~al.(2020)Ali, Jaiswal, Kodge, Agrawal, Chakraborty, and
  Roy]{ali2020imac}
Mustafa Ali, Akhilesh Jaiswal, Sangamesh Kodge, Amogh Agrawal, Indranil
  Chakraborty, and Kaushik Roy.
\newblock Imac: In-memory multi-bit multiplication and accumulation in 6t sram
  array.
\newblock \emph{IEEE Transactions on Circuits and Systems I: Regular Papers},
  67\penalty0 (8):\penalty0 2521--2531, 2020.

\bibitem[Bietti et~al.(2018)Bietti, Mialon, and
  Mairal]{bietti2018regularization}
Alberto Bietti, Gr{\'e}goire Mialon, and Julien Mairal.
\newblock On regularization and robustness of deep neural networks.
\newblock 2018.

\bibitem[Bojarski et~al.(2016)Bojarski, Del~Testa, Dworakowski, Firner, Flepp,
  Goyal, Jackel, Monfort, Muller, Zhang, et~al.]{bojarski2016end}
Mariusz Bojarski, Davide Del~Testa, Daniel Dworakowski, Bernhard Firner, Beat
  Flepp, Prasoon Goyal, Lawrence~D Jackel, Mathew Monfort, Urs Muller, Jiakai
  Zhang, et~al.
\newblock End to end learning for self-driving cars.
\newblock \emph{arXiv preprint arXiv:1604.07316}, 2016.

\bibitem[Buckman et~al.(2018)Buckman, Roy, Raffel, and
  Goodfellow]{buckman2018thermometer}
Jacob Buckman, Aurko Roy, Colin Raffel, and Ian Goodfellow.
\newblock Thermometer encoding: One hot way to resist adversarial examples.
\newblock In \emph{International Conference on Learning Representations}, 2018.

\bibitem[Bulat \& Tzimiropoulos(2021)Bulat and Tzimiropoulos]{bulat2021bit}
Adrian Bulat and Georgios Tzimiropoulos.
\newblock Bit-mixer: Mixed-precision networks with runtime bit-width selection.
\newblock \emph{arXiv preprint arXiv:2103.17267}, 2021.

\bibitem[Chen et~al.(2021)Chen, Cheng, Gan, Yuan, Zhang, and
  Wang]{chen2021chasing}
Tianlong Chen, Yu~Cheng, Zhe Gan, Lu~Yuan, Lei Zhang, and Zhangyang Wang.
\newblock Chasing sparsity in vision transformers: An end-to-end exploration.
\newblock \emph{Advances in Neural Information Processing Systems}, 34, 2021.

\bibitem[Coates et~al.(2011)Coates, Ng, and Lee]{coates2011analysis}
Adam Coates, Andrew Ng, and Honglak Lee.
\newblock An analysis of single-layer networks in unsupervised feature
  learning.
\newblock In \emph{Proceedings of the fourteenth international conference on
  artificial intelligence and statistics}, pp.\  215--223. JMLR Workshop and
  Conference Proceedings, 2011.

\bibitem[Croce \& Hein(2020)Croce and Hein]{croce2020reliable}
Francesco Croce and Matthias Hein.
\newblock Reliable evaluation of adversarial robustness with an ensemble of
  diverse parameter-free attacks.
\newblock In \emph{ICML}, 2020.

\bibitem[De~Vries et~al.(2017)De~Vries, Strub, Mary, Larochelle, Pietquin, and
  Courville]{de2017modulating}
Harm De~Vries, Florian Strub, J{\'e}r{\'e}mie Mary, Hugo Larochelle, Olivier
  Pietquin, and Aaron Courville.
\newblock Modulating early visual processing by language.
\newblock \emph{arXiv preprint arXiv:1707.00683}, 2017.

\bibitem[Han et~al.(2021)Han, Nebelung, Pedersoli, Zimmermann, Schulze-Hagen,
  Ho, Haarburger, Kiessling, Kuhl, Schulz, et~al.]{han2021advancing}
Tianyu Han, Sven Nebelung, Federico Pedersoli, Markus Zimmermann, Maximilian
  Schulze-Hagen, Michael Ho, Christoph Haarburger, Fabian Kiessling, Christiane
  Kuhl, Volkmar Schulz, et~al.
\newblock Advancing diagnostic performance and clinical usability of neural
  networks via adversarial training and dual batch normalization.
\newblock \emph{Nature Communications}, 12\penalty0 (1):\penalty0 1--11, 2021.

\bibitem[Hansen(2015)]{hansen2015tiny}
Lucas Hansen.
\newblock Tiny {ImageNet} challenge submission.
\newblock \emph{CS 231N}, 2015.

\bibitem[He et~al.(2016)He, Zhang, Ren, and Sun]{he2016deep}
Kaiming He, Xiangyu Zhang, Shaoqing Ren, and Jian Sun.
\newblock Deep residual learning for image recognition.
\newblock In \emph{Proceedings of the IEEE conference on computer vision and
  pattern recognition}, pp.\  770--778, 2016.

\bibitem[He et~al.(2018)He, Lin, Liu, Wang, Li, and Han]{he2018amc}
Yihui He, Ji~Lin, Zhijian Liu, Hanrui Wang, Li-Jia Li, and Song Han.
\newblock Amc: Automl for model compression and acceleration on mobile devices.
\newblock In \emph{Proceedings of the European conference on computer vision
  (ECCV)}, pp.\  784--800, 2018.

\bibitem[He et~al.(2019)He, Rakin, and Fan]{he2019parametric}
Zhezhi He, Adnan~Siraj Rakin, and Deliang Fan.
\newblock Parametric noise injection: Trainable randomness to improve deep
  neural network robustness against adversarial attack.
\newblock In \emph{Proceedings of the IEEE Conference on Computer Vision and
  Pattern Recognition}, pp.\  588--597, 2019.

\bibitem[Horowitz(2014)]{horowitz20141}
Mark Horowitz.
\newblock 1.1 computing's energy problem (and what we can do about it).
\newblock In \emph{2014 IEEE International Solid-State Circuits Conference
  Digest of Technical Papers (ISSCC)}, pp.\  10--14. IEEE, 2014.

\bibitem[Hua et~al.(2021)Hua, Zhang, Guo, Zhang, and Suh]{hua2021bullettrain}
Weizhe Hua, Yichi Zhang, Chuan Guo, Zhiru Zhang, and G~Edward Suh.
\newblock Bullettrain: Accelerating robust neural network training via boundary
  example mining.
\newblock \emph{Advances in Neural Information Processing Systems}, 34, 2021.

\bibitem[Huang et~al.(2017)Huang, Chen, Li, Wu, van~der Maaten, and
  Weinberger]{huang2017multi}
Gao Huang, Danlu Chen, Tianhong Li, Felix Wu, Laurens van~der Maaten, and
  Kilian~Q Weinberger.
\newblock Multi-scale dense networks for resource efficient image
  classification.
\newblock \emph{arXiv preprint arXiv:1703.09844}, 2017.

\bibitem[Huang \& Belongie(2017)Huang and Belongie]{huang2017arbitrary}
Xun Huang and Serge Belongie.
\newblock Arbitrary style transfer in real-time with adaptive instance
  normalization.
\newblock In \emph{Proceedings of the IEEE International Conference on Computer
  Vision}, pp.\  1501--1510, 2017.

\bibitem[Kang et~al.(2018)Kang, Lim, Gonugondla, and Shanbhag]{kang2018memory}
Mingu Kang, Sungmin Lim, Sujan Gonugondla, and Naresh~R Shanbhag.
\newblock An in-memory vlsi architecture for convolutional neural networks.
\newblock \emph{IEEE Journal on Emerging and Selected Topics in Circuits and
  Systems}, 8\penalty0 (3):\penalty0 494--505, 2018.

\bibitem[Kaya et~al.(2019)Kaya, Hong, and Dumitras]{kaya2019shallow}
Yigitcan Kaya, Sanghyun Hong, and Tudor Dumitras.
\newblock Shallow-deep networks: Understanding and mitigating network
  overthinking.
\newblock In \emph{International Conference on Machine Learning}, pp.\
  3301--3310. PMLR, 2019.

\bibitem[Krizhevsky et~al.(2009)Krizhevsky, Hinton,
  et~al.]{krizhevsky2009learning}
Alex Krizhevsky, Geoffrey Hinton, et~al.
\newblock Learning multiple layers of features from tiny images.
\newblock 2009.

\bibitem[Kundu \& Sundaresan(2021)Kundu and Sundaresan]{kundu2021attentionlite}
Souvik Kundu and Sairam Sundaresan.
\newblock Attentionlite: Towards efficient self-attention models for vision.
\newblock In \emph{ICASSP 2021-2021 IEEE International Conference on Acoustics,
  Speech and Signal Processing (ICASSP)}, pp.\  2225--2229. IEEE, 2021.

\bibitem[Kundu et~al.(2021)Kundu, Nazemi, Beerel, and Pedram]{kundu2021dnr}
Souvik Kundu, Mahdi Nazemi, Peter~A Beerel, and Massoud Pedram.
\newblock Dnr: A tunable robust pruning framework through dynamic network
  rewiring of dnns.
\newblock In \emph{Proceedings of the 26th Asia and South Pacific Design
  Automation Conference}, pp.\  344--350, 2021.

\bibitem[Lecuyer et~al.(2019)Lecuyer, Atlidakis, Geambasu, Hsu, and
  Jana]{lecuyer2019certified}
Mathias Lecuyer, Vaggelis Atlidakis, Roxana Geambasu, Daniel Hsu, and Suman
  Jana.
\newblock Certified robustness to adversarial examples with differential
  privacy.
\newblock In \emph{2019 IEEE Symposium on Security and Privacy (SP)}, pp.\
  656--672. IEEE, 2019.

\bibitem[Li et~al.(2021)Li, Yuan, Niu, Zhao, Li, Cai, Shen, Zhan, Kong, Jin,
  et~al.]{li2021npas}
Zhengang Li, Geng Yuan, Wei Niu, Pu~Zhao, Yanyu Li, Yuxuan Cai, Xuan Shen,
  Zheng Zhan, Zhenglun Kong, Qing Jin, et~al.
\newblock Npas: A compiler-aware framework of unified network pruning and
  architecture search for beyond real-time mobile acceleration.
\newblock In \emph{Proceedings of the IEEE/CVF Conference on Computer Vision
  and Pattern Recognition}, pp.\  14255--14266, 2021.

\bibitem[Liu et~al.(2018)Liu, Sun, Zhou, Huang, and Darrell]{liu2018rethinking}
Zhuang Liu, Mingjie Sun, Tinghui Zhou, Gao Huang, and Trevor Darrell.
\newblock Rethinking the value of network pruning.
\newblock \emph{arXiv preprint arXiv:1810.05270}, 2018.

\bibitem[Madry et~al.(2017)Madry, Makelov, Schmidt, Tsipras, and
  Vladu]{madry2017towards}
Aleksander Madry, Aleksandar Makelov, Ludwig Schmidt, Dimitris Tsipras, and
  Adrian Vladu.
\newblock Towards deep learning models resistant to adversarial attacks.
\newblock \emph{arXiv preprint arXiv:1706.06083}, 2017.

\bibitem[Meng \& Chen(2017)Meng and Chen]{meng2017magnet}
Dongyu Meng and Hao Chen.
\newblock Magnet: a two-pronged defense against adversarial examples.
\newblock In \emph{Proceedings of the 2017 ACM SIGSAC Conference on Computer
  and Communications Security}, pp.\  135--147, 2017.

\bibitem[Netzer et~al.(2011)Netzer, Wang, Coates, Bissacco, Wu, and
  Ng]{netzer2011reading}
Yuval Netzer, Tao Wang, Adam Coates, Alessandro Bissacco, Bo~Wu, and Andrew~Y
  Ng.
\newblock Reading digits in natural images with unsupervised feature learning.
\newblock 2011.

\bibitem[Paszke et~al.(2017)Paszke, Gross, Chintala, Chanan, Yang, DeVito, Lin,
  Desmaison, Antiga, and Lerer]{paszke2017automatic}
Adam Paszke, Sam Gross, Soumith Chintala, Gregory Chanan, Edward Yang, Zachary
  DeVito, Zeming Lin, Alban Desmaison, Luca Antiga, and Adam Lerer.
\newblock Automatic differentiation in pytorch.
\newblock 2017.

\bibitem[Perez et~al.(2018)Perez, Strub, De~Vries, Dumoulin, and
  Courville]{perez2018film}
Ethan Perez, Florian Strub, Harm De~Vries, Vincent Dumoulin, and Aaron
  Courville.
\newblock Film: Visual reasoning with a general conditioning layer.
\newblock In \emph{Proceedings of the AAAI Conference on Artificial
  Intelligence}, volume~32, 2018.

\bibitem[Samangouei et~al.(2018)Samangouei, Kabkab, and
  Chellappa]{samangouei2018defense}
Pouya Samangouei, Maya Kabkab, and Rama Chellappa.
\newblock Defense-gan: Protecting classifiers against adversarial attacks using
  generative models.
\newblock \emph{arXiv preprint arXiv:1805.06605}, 2018.

\bibitem[Schmidt et~al.(2018)Schmidt, Santurkar, Tsipras, Talwar, and
  Madry]{schmidt2018adversarially}
Ludwig Schmidt, Shibani Santurkar, Dimitris Tsipras, Kunal Talwar, and
  Aleksander Madry.
\newblock Adversarially robust generalization requires more data.
\newblock \emph{arXiv preprint arXiv:1804.11285}, 2018.

\bibitem[Singh et~al.(2019)Singh, Verma, Rai, and Namboodiri]{singh2019hetconv}
Pravendra Singh, Vinay~Kumar Verma, Piyush Rai, and Vinay~P Namboodiri.
\newblock Hetconv: Heterogeneous kernel-based convolutions for deep cnns.
\newblock In \emph{Proceedings of the IEEE/CVF Conference on Computer Vision
  and Pattern Recognition}, pp.\  4835--4844, 2019.

\bibitem[Sun et~al.(2019)Sun, Zhu, and Lin]{sun2019towards}
Ke~Sun, Zhanxing Zhu, and Zhouchen Lin.
\newblock Towards understanding adversarial examples systematically: Exploring
  data size, task and model factors.
\newblock \emph{arXiv preprint arXiv:1902.11019}, 2019.

\bibitem[Szegedy et~al.(2013)Szegedy, Zaremba, Sutskever, Bruna, Erhan,
  Goodfellow, and Fergus]{szegedy2013intriguing}
Christian Szegedy, Wojciech Zaremba, Ilya Sutskever, Joan Bruna, Dumitru Erhan,
  Ian Goodfellow, and Rob Fergus.
\newblock Intriguing properties of neural networks.
\newblock \emph{arXiv preprint arXiv:1312.6199}, 2013.

\bibitem[Teerapittayanon et~al.(2016)Teerapittayanon, McDanel, and
  Kung]{teerapittayanon2016branchynet}
Surat Teerapittayanon, Bradley McDanel, and Hsiang-Tsung Kung.
\newblock Branchynet: Fast inference via early exiting from deep neural
  networks.
\newblock In \emph{2016 23rd International Conference on Pattern Recognition
  (ICPR)}, pp.\  2464--2469. IEEE, 2016.

\bibitem[Tram{\`e}r et~al.(2017)Tram{\`e}r, Kurakin, Papernot, Goodfellow,
  Boneh, and McDaniel]{tramer2017ensemble}
Florian Tram{\`e}r, Alexey Kurakin, Nicolas Papernot, Ian Goodfellow, Dan
  Boneh, and Patrick McDaniel.
\newblock Ensemble adversarial training: Attacks and defenses.
\newblock \emph{arXiv preprint arXiv:1705.07204}, 2017.

\bibitem[Tritschler(2021)]{tritschler_2021}
Charlie Tritschler, Sep 2021.
\newblock URL
  \url{https://www.aboutamazon.com/news/devices/meet-astro-a-home-robot-unlike-any-other}.

\bibitem[Tsipras et~al.(2018)Tsipras, Santurkar, Engstrom, Turner, and
  Madry]{tsipras2018robustness}
Dimitris Tsipras, Shibani Santurkar, Logan Engstrom, Alexander Turner, and
  Aleksander Madry.
\newblock Robustness may be at odds with accuracy.
\newblock \emph{arXiv preprint arXiv:1805.12152}, 2018.

\bibitem[Wang et~al.(2020)Wang, Chen, Gui, Hu, Liu, and Wang]{wang2020once}
Haotao Wang, Tianlong Chen, Shupeng Gui, Ting-Kuei Hu, Ji~Liu, and Zhangyang
  Wang.
\newblock Once-for-all adversarial training: In-situ tradeoff between
  robustness and accuracy for free.
\newblock \emph{arXiv preprint arXiv:2010.11828}, 2020.

\bibitem[Wang et~al.(2018)Wang, Yu, Dou, Darrell, and
  Gonzalez]{wang2018skipnet}
Xin Wang, Fisher Yu, Zi-Yi Dou, Trevor Darrell, and Joseph~E Gonzalez.
\newblock Skipnet: Learning dynamic routing in convolutional networks.
\newblock In \emph{Proceedings of the European Conference on Computer Vision
  (ECCV)}, pp.\  409--424, 2018.

\bibitem[Xie \& Yuille(2019)Xie and Yuille]{xie2019intriguing}
Cihang Xie and Alan Yuille.
\newblock Intriguing properties of adversarial training at scale.
\newblock \emph{arXiv preprint arXiv:1906.03787}, 2019.

\bibitem[Xie et~al.(2020)Xie, Tan, Gong, Wang, Yuille, and
  Le]{xie2020adversarial}
Cihang Xie, Mingxing Tan, Boqing Gong, Jiang Wang, Alan~L Yuille, and Quoc~V
  Le.
\newblock Adversarial examples improve image recognition.
\newblock In \emph{Proceedings of the IEEE/CVF Conference on Computer Vision
  and Pattern Recognition}, pp.\  819--828, 2020.

\bibitem[Yang et~al.(2019)Yang, Wang, Wang, Xu, Liu, and
  Guo]{yang2019controllable}
Shuai Yang, Zhangyang Wang, Zhaowen Wang, Ning Xu, Jiaying Liu, and Zongming
  Guo.
\newblock Controllable artistic text style transfer via shape-matching gan.
\newblock In \emph{Proceedings of the IEEE/CVF International Conference on
  Computer Vision}, pp.\  4442--4451, 2019.

\bibitem[Yu et~al.(2018)Yu, Yang, Xu, Yang, and Huang]{yu2018slimmable}
Jiahui Yu, Linjie Yang, Ning Xu, Jianchao Yang, and Thomas Huang.
\newblock Slimmable neural networks.
\newblock \emph{arXiv preprint arXiv:1812.08928}, 2018.

\bibitem[Zagoruyko \& Komodakis(2016)Zagoruyko and
  Komodakis]{zagoruyko2016wide}
Sergey Zagoruyko and Nikos Komodakis.
\newblock Wide residual networks.
\newblock \emph{arXiv preprint arXiv:1605.07146}, 2016.

\end{thebibliography}
\bibliographystyle{iclr2022_conference}

\end{document}

\section{Details on Hyperparameters and Models}
\label{sec:expt}

\subsection{Hyperparameters}
\label{subsec:hyper}
We trained the models for 200 epochs with a starting learning rate (LR) of $0.1$ and a cosine annealing LR scheduler \cite{loshchilov2016sgdr}. We used the SGD optimizer with a weight decay and momentum value of $5e-4$ and $0.9$, respectively. For CIFAR-10, CIFAR-100, and SVHN  we used a batch-size of $128$ and for STL-10 which has comparatively larger resolution ($96 \times 96$ as image height/width compared to $32 \times 32$ of others), we used a batch-size of $64$. We used half of each image-batch as clean and remaining half as perturbed (using PGD-7 attack). To reproduce the results with OAT and OATS we used the official repository of \cite{wang2020once} and used the $\lambda$ distribution, encoding dimension, sampling scheme as recommended by them.  

\subsection{Models}
We used ResNet34 to evaluate on CIFAR-10 and CIFAR-100, WRN16-8 on SVHN and WRN40-2 on STL10. Table \ref{tab:model_details} shows the parameters and FLOPs count (contributed by the convolutions and FC layers as they attribute to majority of the FLOPs and parameters) details of each model.

\begin{table}[!h]
\caption{Model parameters and FLOPs details.}
\label{tab:model_details}
\begin{center}
\begin{tabular}{|l|l|l|l|l|}
\hline
\multicolumn{1}{|c|}{\bf Model type}  &\multicolumn{2}{c|}{\bf Parameters (M)} &\multicolumn{2}{c|}{\bf FLOPs (G)}\\
 \cline{2-5}
 & \multicolumn{1}{c|}{\bf OAT} & \multicolumn{1}{c|}{\bf FLOAT} & \multicolumn{1}{c|}{\bf OAT} & \multicolumn{1}{c|}{\bf FLOAT}\\
 \hline
ResNet34         & 31.4 & 21.28 & 1.18 & 1.165\\
\hline
WRN40-2             & 3.22 & 2.25 & 3 & 2.96 \\
\hline
WRN16-8             & 15.43 & 10.97 & 1.6 & 1.55\\
\hline
\end{tabular}
\end{center}
\end{table}

\section{More Results and Plots}
\subsection{Trained Scaling Factors}
Here we show the trained noise scaling-factor $\alpha$ for each layer of a FLOAT model. In particular, as shown in Fig. \ref{fig:trained_alpha_vals}, the $\alpha$ values are generally high at the initial layers of the models while their value reduces to near zero at later layers. \cite{he2019parametric} has also observed similar trend in their trained alpha values while training for only robustness improvement.   
\begin{figure}[!h]
    \includegraphics[width=0.99\textwidth]{iclr2022/Figs/trained_alpha_values_supp.png}
  \centering
  \caption{Trained noise scaling factor value (layer-wise) for (a) ResNet34 on CIFAR-10, (b) ResNet34 on CIFAR-100, and (c) WRN40-2 on STL10.}
  \label{fig:trained_alpha_vals}
  \vspace{-2mm}
\end{figure}

\subsection{Alternate Results Plots for FLOAT in Terms of $\lambda_n$ vs. Accuracy}
\begin{figure}[!t]
    \includegraphics[width=0.99\textwidth]{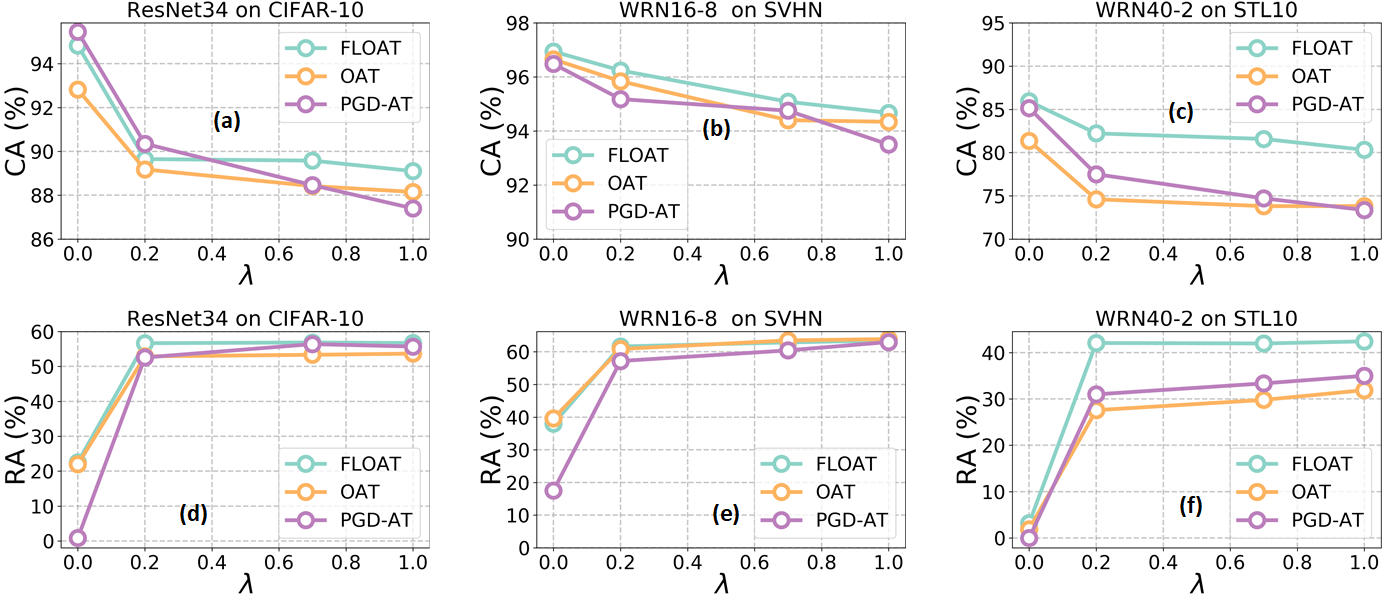}
  \centering
  \caption{Comparison of trade-off between accuracy and robustness of FLOAT, OAT, and PGD-AT. (a)-(c) and (d)-(f) show CA and RA plots vs different $\lambda$ values, respectively.}
  \label{fig:alternate_plots}
  \vspace{-0mm}
\end{figure}

Fig. \ref{fig:alternate_plots} show an alternate plotting of CA-RA comparisons between FLOAT, OAT, and PGD-AT compared to that shown in Fig. 7 of the main manuscript. 

\subsection{Training Time as a Function of Size of $S_{\lambda}$ used for Training.}
Here we show the training time comparison of OAT with different values of training $\lambda$s compared to that for FLOAT. As, for FLOAT, there are always two possible training $\lambda$s, the time remains constant. However, as shown in Fig. \ref{fig:float_oat_ttime_vs_lambda}, for OAT, the training time increases to support more training $\lambda$s. 
In the original manuscript we reported the training time comparisons for OAT training with four training $\lambda$s ({0.0, 0.2, 0.7, 1.0}).
\begin{figure}[!h]
    \includegraphics[width=0.99\textwidth]{iclr2022/Figs/FLOAT_vs_OAT_training_time_vs_lambda_numbers.png}
  \centering
  \caption{Comparison of per epoch normalized training time between FLOAT and OAT for different number of oAT training $\lambda$s, on (a)CIFAR-10, (b) SVHN, and (c) STL10.}
  \label{fig:float_oat_ttime_vs_lambda}
  \vspace{-0mm}
\end{figure}
\bibliography{iclr2022_conference}
\bibliographystyle{iclr2022_conference}